\documentclass[11pt]{article}
\pdfoutput=1 

\usepackage[utf8]{inputenc}
\usepackage[T1]{fontenc}
\usepackage{lmodern} 
\usepackage[margin=1in]{geometry}
\usepackage{amsmath,amssymb}
\usepackage{booktabs}
\usepackage{graphicx}
\usepackage{xcolor}
\usepackage{microtype}
\usepackage[numbers,sort&compress]{natbib}
\usepackage[colorlinks=true,linkcolor=black,citecolor=blue!50!black,urlcolor=blue!50!black]{hyperref}
\usepackage{url}
\usepackage{tikz}
\usetikzlibrary{arrows.meta,positioning,fit,backgrounds,shapes.geometric,calc}
\graphicspath{{figs/}}
\usepackage{listings}
\lstdefinestyle{prompt}{%
  basicstyle=\ttfamily\scriptsize, breaklines=true, breakautoindent=false,
  columns=fullflexible, keepspaces=true, showstringspaces=false,
  frame=single, framesep=3pt, xleftmargin=3pt, xrightmargin=3pt, aboveskip=5pt, belowskip=5pt}

\definecolor{engramblue}{HTML}{1F5C8B}
\definecolor{engramlight}{HTML}{9BB7D4}
\definecolor{baselinered}{HTML}{B0413E}
\definecolor{s2green}{HTML}{2E7D5B}

\newcommand{\engram}{\textsc{Engram}}
\newcommand{\lean}{\texttt{engram\_lean}}
\newcommand{\full}{\texttt{full\_context}}

\title{\bfseries Less Context, More Accuracy:\\
A Bi-Temporal Memory Engine for LLM Agents\\
Where a Lean Retrieved Context Beats the Full History}

\author{%
  Liuyin Wang\thanks{Correspondence: \texttt{liuyinwangthu@gmail.com}. Code, raw logs, and the
  reproducible harness: \url{https://github.com/ly-wang19/engram}.}\\
  Independent Researcher%
}
\date{June 2026}

\begin{document}
\maketitle

\begin{abstract}
Long-term memory is the missing layer for LLM agents: across sessions they forget, and the common
workaround---replaying the entire history into the prompt---is expensive, slow, and, as distractors
accumulate, \emph{less} accurate. Most memory systems win on cost or latency but still lose to the
full-context baseline on accuracy, and the field's benchmark numbers are reported on inconsistent,
non-reproducible harnesses, so the same system appears at wildly different scores across sources.
We present \engram{}, an open-source, dual-process memory engine built on a bi-temporal data model. A fast
write path appends lossless episodes without an LLM on the critical path; an asynchronous consolidation
path extracts atomic $(\textit{subject},\textit{predicate},\textit{object})$ facts, builds a bi-temporal
knowledge graph, and resolves contradictions \emph{without} an LLM call per fact---invalidating, never
deleting, so every fact retains provenance and a supersession chain. A hybrid read path fuses dense, lexical,
graph, and recency/salience signals, applies a point-in-time (``as-of'') temporal filter, and assembles a
compact, provenance-tagged context. On the full 500-question \textsc{LongMemEval}\textsubscript{S} benchmark,
graded by the \emph{official} category-specific judge, \engram{}'s lean configuration---which answers from a
$\sim$9.6k-token retrieved slice, never the full history---scores \textbf{83.6\%} versus \textbf{73.2\%} for
the full-context baseline ($+10.4$ points, McNemar exact $p<10^{-6}$) while using \textbf{$\sim$8$\times$
fewer tokens} (9.6k vs.\ 79k), with 0 of 500 questions errored under both systems. We find the gain is load-bearing on the read path being
\emph{hybrid}: a facts-only path loses recall, while facts plus retrieved raw chunks recover detail. Beyond
the system, we contribute a single neutral, in-repo evaluation harness with the official judge baked in, the
full-context baseline in every table, and the raw per-question logs published---and we document the
measurement-integrity pitfalls (truncation, home-grown judges, full-history ``leaks'') that silently distort
memory-benchmark numbers. Every number in this paper ships with the command to reproduce it.
\end{abstract}

\section{Introduction}

LLM agents are stateless across sessions. The pragmatic fix---concatenate the whole conversation history into
the prompt---scales badly on three axes at once: token cost grows linearly with history, latency follows, and
accuracy \emph{degrades} as irrelevant turns crowd the window and the model is forced to locate a needle among
distractors~\citep{liu2024lost}. A long-term memory layer that stores, structures, and selectively retrieves
the past is the natural alternative, and a growing body of systems pursues
it~\citep{packer2023memgpt,park2023generative,chhikara2025mem0,rasmussen2025zep,gutierrez2024hipporag};
see \citet{du2026memorysurvey} for a recent (2026) survey.

Two gaps remain wide open. First, \textbf{accuracy}: most memory systems are reported as cheaper or faster
than full-context, but \emph{not more accurate}. Beating full-context \emph{on accuracy}---a precisely
retrieved slice that outperforms the noisy full window---is the harder and more valuable target, because it
turns memory from a cost optimization into a quality improvement. Second, \textbf{reproducibility}: memory
benchmarks are reported on inconsistent harnesses with different ingestion, answer prompts, and judges, so a
single system can appear at 58\%, 66\%, or 92\% depending on the source, and different papers give
contradictory orderings. In a field where every number is contested, a neutral harness anyone can re-run is
itself a contribution.

We address both with \engram{}, an open-source memory engine, and its in-repo evaluation harness. Our design
follows a single principle---\emph{a number we cannot reproduce does not exist}---and a composition thesis:
no single mechanism wins, so we compose typed memory, a bi-temporal knowledge graph, multi-signal retrieval
fusion, salience decay, and asynchronous consolidation, and build in the seams between them.

\paragraph{Contributions.}
\begin{enumerate}
  \item \textbf{A dual-process, bi-temporal memory engine} (\S\ref{sec:system}) with cheap, non-destructive
  conflict resolution: a contradicted fact is \emph{invalidated} (with a \texttt{supersedes} chain and full
  provenance), never overwritten, and the common case is resolved with \emph{no} LLM call.
  \item \textbf{The empirical result that a lean retrieved context beats full-context on accuracy}
  (\S\ref{sec:results}): $+10.4$ points (83.6\% vs.\ 73.2\%) at $\sim$8$\times$ fewer tokens on the full
  500-question \textsc{LongMemEval}\textsubscript{S} under the official judge---i.e.\ removing distractors
  \emph{raises} accuracy---together with the finding that a \emph{hybrid} facts-plus-chunks read path is
  load-bearing (facts alone lose recall).
  \item \textbf{A neutral, reproducible harness} (\S\ref{sec:setup}): one in-repo pipeline with the official
  judge baked in, the full-context baseline in every table, the same answerer and judge applied to every
  system by construction, and the raw per-question logs published. We additionally document the
  measurement-integrity pitfalls we found and fixed.
  \item \textbf{A per-category analysis} (\S\ref{sec:results}) isolating where bi-temporal modeling pays off
  (knowledge-update 87.5\%, temporal 81.1\%) and where headroom remains (multi-session aggregation,
  preference).
\end{enumerate}

\engram{} is dual-licensed (AGPL-3.0 plus a commercial license), self-hostable, and runs end-to-end with zero
setup---no API keys, no external services---via deterministic offline fallbacks, so the architecture and the
demo are reproducible without network access.

\section{Related Work}

\paragraph{Memory systems for LLM agents.}
MemGPT~\citep{packer2023memgpt} frames the LLM as an operating system that pages memory in and out of a
limited context window. Generative Agents~\citep{park2023generative} introduce a memory stream with
importance, recency, and relevance scoring plus periodic reflection. Mem0~\citep{chhikara2025mem0} targets
production agents with scalable extract-and-store memory. Zep/Graphiti~\citep{rasmussen2025zep} is the closest
in spirit to \engram{}: a temporally-aware knowledge-graph memory with bi-temporal edges.
HippoRAG~\citep{gutierrez2024hipporag} uses a graph plus personalized PageRank for multi-hop retrieval, and
its successor HippoRAG~2~\citep{gutierrez2025hipporag2} reframes the same machinery as non-parametric continual
memory. The most recent agentic-memory systems add LLM-driven control over how memories are written and
organized: A-MEM~\citep{xu2025amem} builds an interlinked, Zettelkasten-style note network;
MemoryOS~\citep{kang2025memoryos} imposes an operating-system-style short/mid/long-term hierarchy; and
GAM~\citep{wu2026gam} decouples memory encoding from consolidation over hierarchical graphs. \engram{} differs in
\emph{composing} a bi-temporal graph with a hybrid (facts + raw chunks) read path and an explicit, cheap
conflict-resolution policy, and---distinctively---in shipping the neutral harness that lets these design
choices be measured rather than asserted.

\paragraph{Long context and the cost of distractors.}
``Lost in the middle''~\citep{liu2024lost} shows that LLMs use long contexts unevenly and that accuracy
degrades when relevant evidence is buried among distractors. This is the mechanism our headline result
exploits: a filtered, precisely retrieved slice can be \emph{more} accurate than the full window, not merely
cheaper.

\paragraph{Benchmarks.}
\textsc{LongMemEval}~\citep{wu2025longmemeval} evaluates chat assistants on long-term interactive memory
across categories (single/multi-session, knowledge-update, temporal reasoning, preference, and abstention),
with category-specific judge prompts. LOCOMO~\citep{maharana2024locomo} evaluates very long-term
conversational memory, and more recent benchmarks extend the setting to multimodal histories
(Mem-Gallery~\citep{bei2026memgallery}). We report on \textsc{LongMemEval}\textsubscript{S} here and treat
LOCOMO and additional backbones as immediate future work (\S\ref{sec:limitations}).

\paragraph{Retrieval.}
\engram{}'s read path builds on dense retrieval~\citep{karpukhin2020dpr} with modern dense sentence
embeddings (the BGE family~\citep{xiao2024cpack}; we use the English \texttt{bge-small-en-v1.5}), classical
lexical scoring (BM25~\citep{robertson2009bm25}), and rank fusion
via Reciprocal Rank Fusion~\citep{cormack2009rrf}, in the retrieval-augmented generation
tradition~\citep{lewis2020rag}. The dual-process framing draws on the System-1/System-2 distinction in
cognitive science~\citep{kahneman2011thinking}, and salience decay on the classical forgetting
curve~\citep{ebbinghaus1913memory}.

\section{The \engram{} System}
\label{sec:system}

\begin{figure}[t]
\centering
\resizebox{\textwidth}{!}{%
\begin{tikzpicture}[
  font=\footnotesize,
  chip/.style={rounded corners=2pt, draw=black!55, fill=white, inner sep=3pt,
               minimum height=8.5mm, align=center},
  store/.style={cylinder, shape border rotate=90, aspect=0.28, draw=engramblue!70,
                fill=engramblue!10, minimum height=11mm, minimum width=22mm, align=center,
                inner sep=1pt},
  io/.style={font=\itshape, text=black!75, align=center},
  ttl/.style={font=\bfseries, anchor=west},
  >={Stealth[length=2.6mm]},
  flow/.style={->, line width=1pt, draw=black!75},
  vflow/.style={->, line width=1.2pt, draw=black!60},
]
\node[io] (in) at (3.0,10.2) {\texttt{add(messages)}};
\node[ttl, text=engramblue] (t1) at (-0.3,9.3) {SYSTEM-1 \ $\cdot$\ hot write path \ $\cdot$\ no LLM \ $\cdot$\ $<$50\,ms};
\node[chip] (s1a) at (0.3,8.4) {append lossless\\Episode};
\node[chip, right=6mm of s1a] (s1b) {identity resolution\\(sessions/devices)};
\node[chip, right=6mm of s1b] (s1c) {light embed\\$+$ enqueue};
\draw[flow] (s1a)--(s1b); \draw[flow] (s1b)--(s1c);
\node[ttl, text=s2green] (t2) at (-0.3,6.9) {SYSTEM-2 \ $\cdot$\ async consolidation \ $\cdot$\ seconds};
\node[chip] (s2a) at (0.3,6.0) {extract atomic\\Facts $(s,p,o)$};
\node[chip, right=6mm of s2a] (s2b) {build bi-temporal\\knowledge graph};
\node[chip, right=6mm of s2b] (s2c) {conflict detect\\$\to$ invalidate};
\node[chip, right=6mm of s2c] (s2d) {salience\\$+$ decay};
\draw[flow] (s2a)--(s2b); \draw[flow] (s2b)--(s2c); \draw[flow] (s2c)--(s2d);
\node[ttl] (t3) at (-0.3,4.4) {TYPED MEMORY \ $\cdot$\ pluggable stores};
\node[store] (m1) at (1.6,3.3) {Episodic};
\node[store, right=8mm of m1] (m2) {Semantic graph\\(bi-temporal)};
\node[store, right=8mm of m2] (m3) {Profile /\\Identity};
\node[store, right=8mm of m3] (m4) {Procedural};
\node[ttl, text=baselinered] (t4) at (-0.3,1.9) {READ PATH \ $\cdot$\ hybrid retrieval \ $\cdot$\ $<$100\,ms};
\node[chip] (r1) at (0.3,1.1) {query\\decompose\\(multi-hop)};
\node[chip, right=4.5mm of r1] (r2) {dense $+$ BM25\\$+$ graph $+$ recency};
\node[chip, right=4.5mm of r2] (r3) {RRF fusion\\($+$ rerank)};
\node[chip, right=4.5mm of r3] (r4) {bi-temporal\\as-of filter};
\node[chip, right=4.5mm of r4] (r5) {abstention\\gate};
\node[chip, right=4.5mm of r5] (r6) {assemble\\context};
\draw[flow] (r1)--(r2); \draw[flow] (r2)--(r3); \draw[flow] (r3)--(r4);
\draw[flow] (r4)--(r5); \draw[flow] (r5)--(r6);
\node[io, left=6mm of r1] (q) {\texttt{search}\\\texttt{(query)}};
\draw[flow] (q)--(r1);
\node[io, right=6mm of r6, text=engramblue] (out) {answer-ready\\context};
\draw[flow] (r6)--(out);
\begin{scope}[on background layer]
\node[rounded corners=4pt, fill=engramblue!6, draw=engramblue!45, fit=(t1)(s1a)(s1c), inner sep=5pt] (bandS1) {};
\node[rounded corners=4pt, fill=s2green!7, draw=s2green!45, fit=(t2)(s2a)(s2d), inner sep=5pt] (bandS2) {};
\node[rounded corners=4pt, fill=black!4, draw=black!35, fit=(t3)(m1)(m4), inner sep=5pt] (bandM) {};
\node[rounded corners=4pt, fill=baselinered!5, draw=baselinered!45, fit=(t4)(r1)(r6), inner sep=5pt] (bandR) {};
\end{scope}
\draw[vflow] (in.south)--(bandS1.north);
\draw[vflow] (bandS1.south)--node[right=1pt,font=\scriptsize,text=black!60]{async queue}(bandS2.north);
\draw[vflow] (bandS2.south)--node[right=1pt,font=\scriptsize,text=black!60]{write facts}(bandM.north);
\draw[vflow] (bandM.south)--node[right=1pt,font=\scriptsize,text=black!60]{retrieve}(bandR.north);
\end{tikzpicture}%
}
\caption{The \engram{} dual-process architecture. A hot write path (\textsc{System-1}) never blocks on an
LLM; an asynchronous consolidation path (\textsc{System-2}) extracts atomic facts, builds the bi-temporal
knowledge graph, and resolves conflicts non-destructively; both feed a typed, bi-temporal memory backed by
pluggable stores; and a hybrid read path retrieves a compact, provenance-tagged slice (dense $+$ lexical $+$
graph $+$ recency, fused by RRF), applies an as-of temporal filter and an abstention gate, and assembles the
answer context.}
\label{fig:arch}
\end{figure}
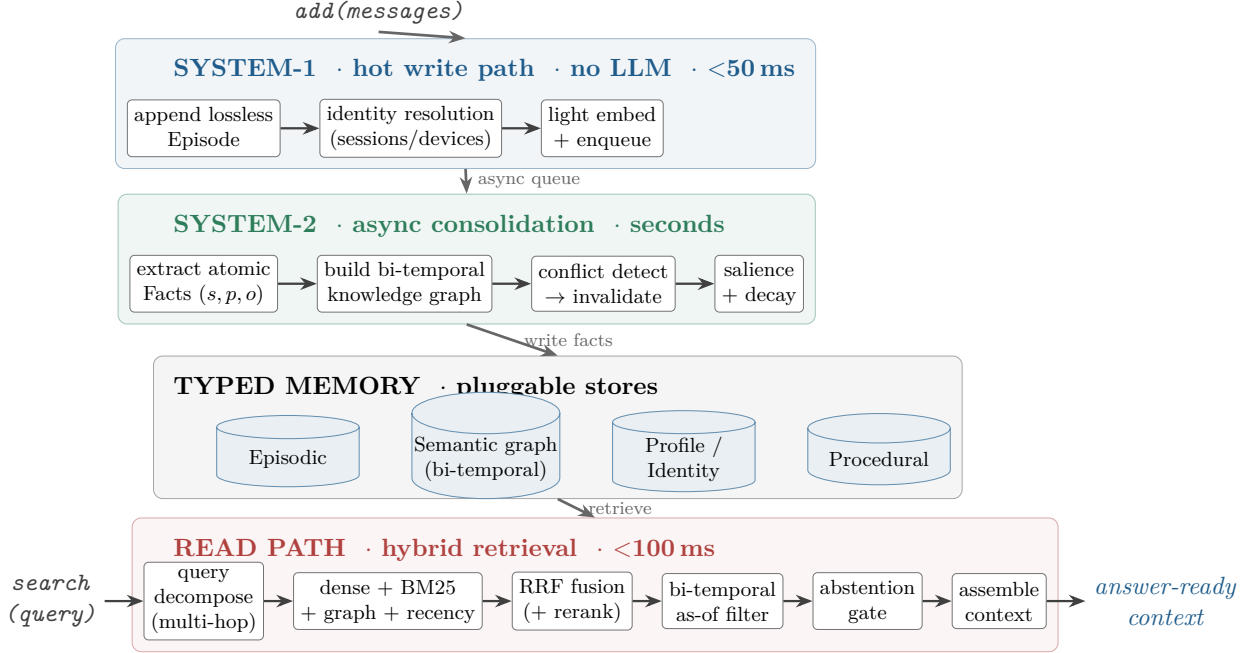

\engram{} is a dual-process memory system: a fast online write path (System-1) and a slow asynchronous
consolidation path (System-2) that writes into a typed, bi-temporal memory, which a hybrid read path queries
(Figure~\ref{fig:arch}).

\subsection{Bi-temporal data model}

Two time axes are first-class everywhere, which is what makes knowledge-updates and ``as-of'' queries
intrinsic rather than bolted-on.

\begin{itemize}
  \item \textbf{Episode} --- a raw, lossless turn/event, stamped with \emph{event time} (when it happened in
  the world) and \emph{ingested-at} (transaction time, when we recorded it).
  \item \textbf{Fact} --- an atomic $(\textit{subject},\textit{predicate},\textit{object})$ claim with surface
  text, an embedding, a \emph{salience} and \emph{confidence}, and \emph{provenance} (the source episode
  ids). It carries two time axes: \emph{valid time} (\texttt{valid\_at} / \texttt{invalid\_at}, when the claim
  is true in the world) and \emph{transaction time} (\texttt{created\_at} / \texttt{expired\_at}, when we
  learned or retracted it), plus a \texttt{supersedes} pointer to the fact it replaces.
  \item \textbf{Entity / Relation} --- graph nodes and edges; edges carry the same bi-temporal stamps.
\end{itemize}

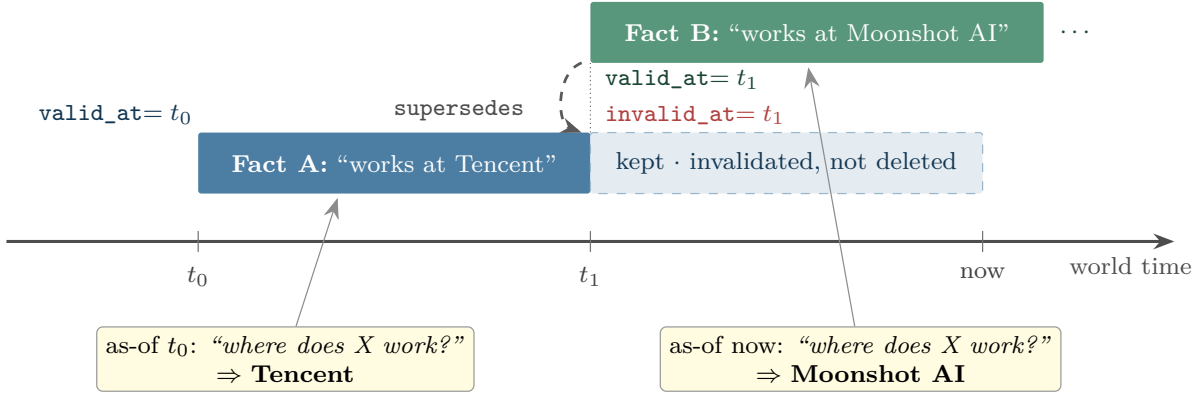
\begin{figure}[t]
\centering
\resizebox{0.96\textwidth}{!}{%
\begin{tikzpicture}[font=\footnotesize, >={Stealth[length=2.4mm]}]
\draw[->, line width=1pt, black!70] (-0.2,0)--(13.2,0);
\node[font=\scriptsize, text=black!70, below] at (12.7,-0.04) {world time};
\foreach \x/\lab in {2/{$t_0$}, 6.5/{$t_1$}, 11/{now}}{
  \draw[black!55] (\x,0.12)--(\x,-0.12);
  \node[below, font=\scriptsize, text=black!75] at (\x,-0.16) {\lab};
}
\fill[engramblue!80, rounded corners=1pt] (2,0.55) rectangle (6.5,1.25);
\node[white, font=\scriptsize] at (4.25,0.9) {\textbf{Fact A:} ``works at Tencent''};
\draw[fill=engramblue!12, draw=engramblue!45, dashed, rounded corners=1pt] (6.5,0.55) rectangle (11,1.25);
\node[text=engramblue!60!black, font=\scriptsize] at (8.75,0.9) {kept $\cdot$ invalidated, not deleted};
\node[font=\scriptsize, text=engramblue!60!black, above left=-1pt and -2pt] at (2,1.25) {\texttt{valid\_at}$=t_0$};
\fill[s2green!80, rounded corners=1pt] (6.5,2.05) rectangle (11.7,2.75);
\node[white, font=\scriptsize] at (9.1,2.4) {\textbf{Fact B:} ``works at Moonshot AI''};
\node[text=s2green!55!black, font=\scriptsize, right] at (11.75,2.4) {$\cdots$};
\draw[densely dotted, black!55] (6.5,1.25)--(6.5,2.05);
\node[font=\scriptsize, text=baselinered, right=1pt] at (6.5,1.45) {\texttt{invalid\_at}$=t_1$};
\node[font=\scriptsize, text=s2green!55!black, right=1pt] at (6.5,1.86) {\texttt{valid\_at}$=t_1$};
\draw[->, dashed, line width=0.9pt, black!65] (6.4,2.05) to[out=-160,in=160] (6.4,1.25);
\node[font=\scriptsize, text=black!70, left=5pt] at (6.05,1.5) {\texttt{supersedes}};
\node[rounded corners=2pt, draw=black!40, fill=yellow!14, font=\scriptsize, align=center, inner sep=3pt]
  (qa) at (3,-1.35) {as-of $t_0$: \emph{``where does X work?''}\\$\Rightarrow$ \textbf{Tencent}};
\node[rounded corners=2pt, draw=black!40, fill=yellow!14, font=\scriptsize, align=center, inner sep=3pt]
  (qb) at (9.6,-1.35) {as-of now: \emph{``where does X work?''}\\$\Rightarrow$ \textbf{Moonshot AI}};
\draw[->, black!45] (qa)--(3.6,0.5);
\draw[->, black!45] (qb)--(9.0,2.0);
\end{tikzpicture}%
}
\caption{Bi-temporal facts make contradictions and ``as-of'' queries first-class. When Fact~B arrives at
$t_1$, \engram{} sets $\text{A.}\texttt{invalid\_at}=t_1$ and $\text{B.}\texttt{supersedes}=\text{A}$: Fact~A
is \emph{invalidated, not deleted}, so the history (and provenance) survives. A point-in-time query then
resolves against the valid fact for that time---\textsc{Tencent} as-of $t_0$, \textsc{Moonshot~AI} as-of
now---which is exactly what the knowledge-update and temporal categories require.}
\label{fig:bitemporal}
\end{figure}

The invariant: \emph{never hard-delete a contradicted fact---invalidate it} (set \texttt{invalid\_at}) and
keep the history. Every fact can therefore answer ``where did this come from?'' and ``what did it replace?''
(Figure~\ref{fig:bitemporal}).

\subsection{System-1: the hot write path}

On \texttt{add(messages)}, \engram{} appends a lossless episode, resolves identity (linking a user/entity
across sessions and devices), computes a light embedding, and enqueues the episode for consolidation. No LLM
runs on this path, keeping it within a sub-50ms budget so writes never block the agent.

\subsection{System-2: asynchronous consolidation}

Off the critical path, the consolidation engine (i) extracts atomic facts from episodes (rule-based by
default, LLM-based when a model is configured), (ii) builds the bi-temporal knowledge graph of entities and
relations, (iii) detects conflicts and invalidates superseded facts, and (iv) scores salience and applies
decay/reinforcement so unreinforced memories fade and the store stays small and fast. Hierarchical
abstraction (session summaries $\rightarrow$ profile) runs here as well.

\subsection{Cheap-then-escalate conflict resolution}

When a new fact arrives for an existing $(\textit{subject},\textit{predicate})$ slot with a different object,
\engram{} resolves it in increasing order of cost:
\begin{enumerate}
  \item \textbf{Slot match} (exact) signals a likely update; \textbf{embedding similarity} catches the same
  attribute under a different free-form predicate; \textbf{content subsumption} (one claim $\subseteq$ the
  other) separates contradiction from elaboration.
  \item If the claims are clearly contradictory and temporally ordered, invalidate the old one
  ($\texttt{old.invalid\_at} \leftarrow \texttt{new.valid\_at}$) and set
  $\texttt{new.supersedes} \leftarrow \texttt{old.id}$. \emph{No LLM call.}
  \item Only genuinely ambiguous cases escalate to an LLM adjudicator.
\end{enumerate}
This is the cost win over systems that invoke an LLM on every fact, while preserving production-grade temporal
correctness and a complete audit trail.

\subsection{The hybrid read path}

On \texttt{search(query)}, \engram{} (1) understands the query and, for multi-hop questions, decomposes it
into sub-queries; (2) retrieves in parallel through four complementary channels---dense semantic, BM25
lexical, graph $n$-hop from the query's entities, and recency/salience; (3) fuses the ranked lists with
Reciprocal Rank Fusion~\citep{cormack2009rrf} and an optional cross-encoder rerank over the top-$k$ (off by
default); (4) applies a bi-temporal ``as-of'' filter (what we believed was true at time $T$); (5) passes an
abstention gate that declines to answer when the evidence is absent; and (6) assembles a deduplicated,
provenance-tagged, token-budgeted context. The per-item score combines all signals:
\begin{equation}
\label{eq:score}
\begin{split}
\mathrm{score}(\textit{item} \mid q) ={}&
  w_{\text{sem}}\cos(q,\textit{item})
  + w_{\text{lex}}\,\mathrm{bm25}(q,\textit{item})
  + w_{\text{graph}}\,\mathrm{prox}(\textit{item}) \\
  &{}+ w_{\text{rec}}\,e^{-\Delta t/\tau}
  + w_{\text{sal}}\,\mathrm{sal}(\textit{item}),
\end{split}
\end{equation}
where the weights are configuration-driven and tuned on the harness rather than hand-set. Crucially, the
assembled context is \emph{hybrid}: it contains both the conflict-resolved bi-temporal facts and the most
relevant raw session chunks (plus session-level summaries). \S\ref{sec:results} shows this hybrid composition
is necessary---facts alone lose recall.

\paragraph{Pluggable backends.}
Every external dependency---LLM, embedder, vector store, graph store, lexical index---sits behind an interface
with a zero-dependency offline fallback (a hashing embedder, a rule-based extractor, in-memory stores). The
end-to-end loop and the unit tests therefore run deterministically with no API keys and no services; real
backends (BGE embeddings, LanceDB/Qdrant/pgvector, Kuzu/Neo4j, any LLM via LiteLLM) slot in behind the same
interfaces.

\section{Experimental Setup}
\label{sec:setup}

\paragraph{Benchmark.}
\textsc{LongMemEval}\textsubscript{S}~\citep{wu2025longmemeval}: 500 questions, each over a haystack of
$\sim$50 sessions ($\sim$115k tokens), pulled from the public release. Questions span seven categories
(Figure~\ref{fig:percat}). We grade with the \emph{official} category-specific judge prompts---including the
temporal off-by-one tolerance, knowledge-update old-information tolerance, preference-rubric leniency, the
``contains the answer'' semantics, and unanswerable (abstention) detection---rather than a home-grown judge.

\paragraph{Models.}
Embedder: \texttt{BAAI/bge-small-en-v1.5} (local, no API key). System-2 fact extractor:
\texttt{doubao-seed-1.6-flash}. Answerer: \texttt{doubao-seed-2.0-pro}. Judge: \texttt{DeepSeek-V3.2}, a
strict standard judge. Models are addressed as \texttt{provider:model} and resolved via LiteLLM, so any
OpenAI-compatible endpoint (OpenAI, DeepSeek, a local model) can be substituted and the same commands run.

\paragraph{Systems.}
\lean{} (our headline) retrieves a small hybrid slice---conflict-resolved bi-temporal facts $+$ the most
relevant raw session chunks $+$ session summaries---and answers from \emph{that alone} ($\sim$9.6k tokens),
never the full history. \full{} is the baseline: stuff the entire haystack into the prompt ($\sim$79k tokens).
The harness applies the \emph{same answerer and judge to every system}, so any within-run comparison is
apples-to-apples by construction.

\paragraph{Retrieval configuration.}
The hybrid read path fuses five ranked signals by weighted Reciprocal Rank Fusion ($k_{\text{RRF}}=60$) with
the weights in Table~\ref{tab:hyper} (repository defaults, tuned on the harness, not hand-set per question).
The lean slice assembles up to 8 conflict-resolved facts, the top-15 fused items, 2 raw session chunks, and
28 session-level summaries; exact flag semantics are in the reproduce command below.

\begin{table}[h]
\centering\small
\caption{Retrieval-fusion weights for Eq.~\eqref{eq:score} and the RRF constant, as used by the headline
\lean{} configuration (defaults in \texttt{engram/config.py}).}
\label{tab:hyper}
\begin{tabular}{cccccc}
\toprule
$w_{\text{sem}}$ & $w_{\text{lex}}$ & $w_{\text{graph}}$ & $w_{\text{rec}}$ & $w_{\text{sal}}$ & $k_{\text{RRF}}$ \\
\midrule
1.0 & 0.6 & 0.8 & 0.3 & 0.25 & 60 \\
\bottomrule
\end{tabular}
\end{table}

\paragraph{Reproduce.}
The headline run is a single command (raw per-question logs are committed to the repository):
\begin{quote}\small\ttfamily
python eval/bench.py --data s --limit 500 \textbackslash\\
\hspace*{2em}--systems engram\_lean,full\_context \textbackslash\\
\hspace*{2em}--answerer volcano:doubao-seed-2-0-pro \textbackslash\\
\hspace*{2em}--judge volcano:deepseek-v3-2 --extractor volcano:doubao-seed-1-6-flash \textbackslash\\
\hspace*{2em}--embedder bge-small --reasoning --persona \textbackslash\\
\hspace*{2em}--chunks 2 --topk 15 --extract-k 8 --summ-k 28 --n-summaries 28
\end{quote}
\texttt{python eval/report.py <run.jsonl>} recomputes every table below from the logs.

\paragraph{Measurement-integrity notes.}
We document the bugs that silently inflate or deflate memory-benchmark numbers, because they are the reason
cross-source numbers disagree:
\begin{enumerate}
  \item \textbf{Lean, not full-history (the honest test).} An earlier headline prepended facts \emph{above the
  entire history}; because that system \emph{contains} full-context, it cannot really lose to it and does not
  validate the retrieval thesis. The headline is now \lean{}, which answers from a $\sim$9.6k-token retrieved
  slice.
  \item \textbf{Full-context truncation.} The baseline was once capped below the haystack size, feeding it
  only the oldest sessions; this \emph{deflated the baseline}. Fixed so it receives the whole haystack. (Any
  ``full-context only scores 30\%'' claim predates this fix.)
  \item \textbf{Official judge, not a home-grown one.} A generic ``same info?'' judge was \emph{stricter} than
  the official LongMemEval judge and made scores non-comparable; we use the official prompts verbatim.
  \item \textbf{Abstention} questions are graded by the official \emph{unanswerable} judge.
  \item \textbf{Reliability.} The LLM client uses exponential-backoff retry with transient/permanent error
  classification; the headline run completed with \textbf{0 errored questions} of 500 under both systems.
\end{enumerate}

\section{Results}
\label{sec:results}

\begin{figure}[t]
\centering
\includegraphics[width=0.72\textwidth]{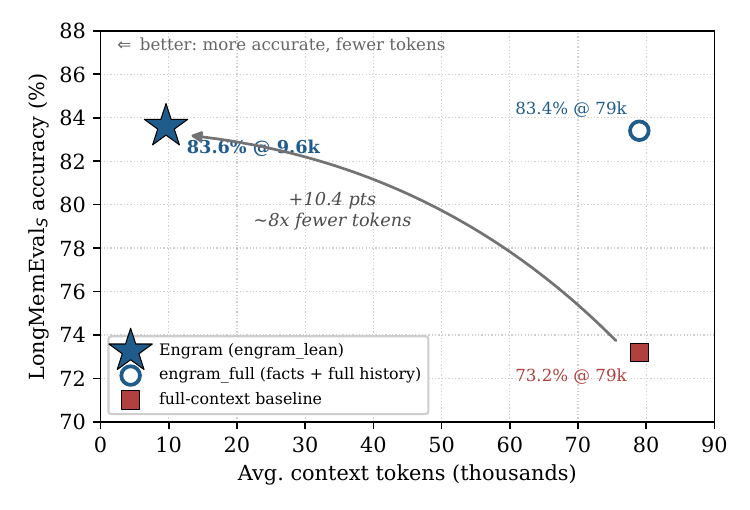}
\caption{Accuracy vs.\ average context tokens on LongMemEval\textsubscript{S} (500 questions, official judge).
\lean{} (star) sits up and to the left of the full-context baseline (square)---\textbf{$+10.4$ points at
$\sim$8$\times$ fewer tokens}. The \texttt{engram\_full} variant (open circle) prepends the same facts above
the \emph{whole} history and lands at 83.4\%: the structured facts carry the accuracy, while the full history
adds tokens, not correctness.}
\label{fig:acctokens}
\end{figure}

\paragraph{Headline.}
Table~\ref{tab:headline} and Figure~\ref{fig:acctokens} report the full 500-question result. \engram{}'s lean configuration beats the
full-context baseline by \textbf{$+10.4$ points} (83.6\% vs.\ 73.2\%) while using \textbf{$\sim$8$\times$
fewer tokens} (9.6k vs.\ 79k), with 0 of 500 errored under both. The filtered slice is not merely cheaper---it
is \emph{more accurate} than the full window, consistent with the distractor mechanism of \citet{liu2024lost}.
And because the retrieved slice is bounded, \engram{}'s cost stays flat as history grows, whereas
full-context cannot. The margin is statistically decisive: across the 500 paired questions \lean{} is correct
on 81 that the baseline misses versus 29 the other way (McNemar's exact test, $p<10^{-6}$), and a paired
bootstrap puts the 95\% CI of the gain at $[+6.4,+14.4]$ points. All statistics are recomputed from the
committed logs by \texttt{paper/compute\_stats.py}.

\begin{table}[h]
\centering
\caption{\textsc{LongMemEval}\textsubscript{S}, 500 questions, official judge. Same answerer
(\texttt{doubao-seed-2.0-pro}) and judge (\texttt{DeepSeek-V3.2}) applied to both systems. Accuracy is shown
with a Wilson 95\% confidence interval; the paired difference is significant (McNemar exact $p<10^{-6}$).}
\label{tab:headline}
\begin{tabular}{lcrr}
\toprule
System & Overall acc.\ (95\% CI) & Avg.\ context tokens & Errors \\
\midrule
\textbf{\engram{}} (\lean{}) & \textbf{83.6\%} \,[80.1,\,86.6] & \textbf{9.6k} & 0 / 500 \\
full-context baseline        & 73.2\% \,[69.2,\,76.9]          & 79k           & 0 / 500 \\
\bottomrule
\end{tabular}
\end{table}

\paragraph{Lean retrieval matches full-history accuracy at 1/8 the cost.}
For reference, in the same 500-question run a non-lean variant that prepends the conflict-resolved facts
\emph{above the full history} (\texttt{engram\_full}, $\sim$79k tokens) scores 83.4\% (416/499; one question
errored under this variant, versus none for \lean{} and full-context). A paired McNemar test against \lean{}'s
83.6\% finds \emph{no} difference ($p=0.91$): the structured facts contribute essentially all of the accuracy
gain, while the full history adds tokens, not correctness. We therefore headline the lean number.

\begin{figure}[t]
\centering
\includegraphics[width=0.82\textwidth]{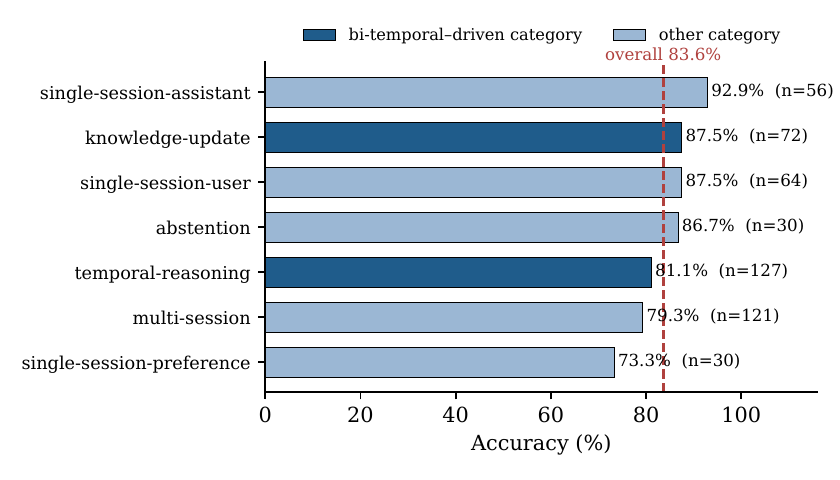}
\caption{Per-category accuracy of \lean{} on the full 500-question set (with per-category $n$; dashed line is
the 83.6\% overall). The two categories where bi-temporal modelling is decisive---\emph{knowledge-update} and
\emph{temporal-reasoning}---are highlighted. Headroom concentrates in \emph{multi-session} aggregation and
\emph{single-session-preference} (hard field-wide).}
\label{fig:percat}
\end{figure}

\paragraph{Per-category.}
Figure~\ref{fig:percat} breaks \lean{} down by category. Bi-temporal modeling pays off where it should:
\emph{knowledge-update} (most-recent-wins via invalidation) reaches 87.5\% and \emph{temporal-reasoning}
(date-stamped context $+$ as-of filtering) 81.1\%. \emph{Abstention} reaches 86.7\% under the official
unanswerable judge---the system declines when memory lacks the answer rather than hallucinating. The headroom
is concentrated in \emph{multi-session} aggregation (counting/aggregating across many sessions, 79.3\%) and
\emph{single-session-preference} (73.3\%, a category that is hard field-wide).

\paragraph{The read path must be hybrid.}
A load-bearing finding from development: a \emph{facts-only} read path---answering purely from extracted
$(s,p,o)$ facts---\emph{loses recall} relative to the hybrid path, because extraction drops detail that some
questions need verbatim. Adding the most relevant raw session chunks back alongside the conflict-resolved
facts restores that detail; the facts contribute conflict-resolved, bi-temporal signal and the chunks restore
specificity. The headline configuration is hybrid for exactly this reason, and we caution against shipping
facts-only QA. We report this as a design observation; a controlled facts-only ablation on the full
500-question set under the same judge is reported as future work (\S\ref{sec:limitations}), not claimed here.

\paragraph{Efficiency.}
The retrieved context averages $\sim$9.6k tokens, $\sim$8$\times$ leaner than the $\sim$79k full-context
baseline, and the lean read path is what keeps cost flat as history grows. Retrieval itself is sub-second; the
$\sim$60s p50 end-to-end latency is dominated by the answerer model's generation call, not by \engram{}.

\section{Discussion and Limitations}
\label{sec:limitations}

We report the result openly rather than as a leaderboard ``win,'' precisely because our thesis is that
cross-harness numbers are not comparable. The honest scope of the present evidence:

\begin{itemize}
  \item \textbf{One benchmark, one answerer backbone.} The headline is \textsc{LongMemEval}\textsubscript{S}
  with a single answerer model. The evaluation discipline we advocate requires multiple backbones (a small
  open model \emph{and} a frontier model) and multiple benchmarks; extending to LOCOMO~\citep{maharana2024locomo}
  and a second backbone is immediate next work, so that memory quality is shown not to depend on one model's
  ability to read our structures.
  \item \textbf{Small samples mislead.} During development an 18-item slice once read 83\% when the full-set
  truth was $\sim$58\%; we therefore report \emph{only} full-500 numbers, and every number in this paper is a
  full-set number with committed logs.
  \item \textbf{Open categories.} Multi-session aggregation and preference are where headroom remains; the
  multi-hop query planner and tuned bi-temporal conflict resolution are the levers we expect to move them.
  \item \textbf{Single run; no controlled component ablation yet.} We report one full-set run per system and
  do not yet quantify run-to-run variance from answerer stochasticity, nor a controlled full-set ablation of
  the read path's components (notably the facts-only vs.\ hybrid comparison of \S\ref{sec:results}, which we
  state as an observation). Repeated-run variance and a component ablation are immediate next work.
\end{itemize}

\section{Conclusion}

\engram{} shows that a precisely retrieved, bi-temporal, hybrid context can beat the full-context baseline
\emph{on accuracy}---$+10.4$ points at $\sim$8$\times$ fewer tokens on the full \textsc{LongMemEval}\textsubscript{S}
under the official judge---turning long-term memory from a cost optimization into a quality improvement. Equally,
we contribute the neutral, reproducible harness that makes such a claim checkable: the official judge baked in,
the full-context baseline in every table, documented measurement-integrity pitfalls, and the raw logs published.
In a field where every number is contested, the trustworthy scoreboard is itself the result. The system, the
harness, and the logs are open source.

\section*{Ethics and Broader Impact}

A long-term memory layer necessarily persists personal data across sessions, which raises real privacy
obligations. Two of \engram{}'s core design choices are mitigations as much as features: \emph{non-destructive
invalidation with full provenance} yields an auditable trail of what was believed, when, and from which
source, which makes targeted deletion and ``right to be forgotten'' requests tractable rather than
best-effort; and \emph{salience decay} lets unreinforced personal details fade by default. Because every
component has a zero-dependency offline fallback, \engram{} can run fully locally---no user data need leave the
operator's machine---and the AGPL-3.0 license keeps self-hosted data under the operator's control. The
principal risks are the obverse of the benefits: a memory store concentrates sensitive information and must be
secured, access-controlled, and scoped to genuine consent; and a memory that confidently surfaces a
\emph{stale} or \emph{wrong} fact can mislead, which is precisely why the bi-temporal model and the abstention
gate (decline when memory lacks the answer) are first-class rather than optional.

\section*{Reproducibility Statement}

Every number in this paper is produced by the in-repo harness and backed by committed per-question logs
(prediction, gold, correctness, tokens, and latency for all 500 questions). The end-to-end loop and unit
tests run with no API keys and no external services via deterministic offline fallbacks; the benchmark
numbers require an OpenAI-compatible model endpoint for the answerer, extractor, and judge, all configurable.
The exact command, model identifiers, embedder, and hyper-parameters are in \S\ref{sec:setup}; raw logs and
the harness are in the repository. The significance tests and confidence intervals in \S\ref{sec:results} are
recomputed from the committed logs by \texttt{paper/compute\_stats.py}, with no model calls.

\bibliographystyle{plainnat}
\bibliography{references}

\begin{thebibliography}{21}
\providecommand{\natexlab}[1]{#1}
\providecommand{\url}[1]{\texttt{#1}}
\expandafter\ifx\csname urlstyle\endcsname\relax
  \providecommand{\doi}[1]{doi: #1}\else
  \providecommand{\doi}{doi: \begingroup \urlstyle{rm}\Url}\fi

\bibitem[Bei et~al.(2026)Bei, Wei, Ning, Zhao, Liu, Lin, Zhu, Hamann, He, and
  Tong]{bei2026memgallery}
Yuanchen Bei, Tianxin Wei, Xuying Ning, Yanjun Zhao, Zhining Liu, Xiao Lin,
  Yada Zhu, Hendrik Hamann, Jingrui He, and Hanghang Tong.
\newblock {Mem-Gallery}: Benchmarking multimodal long-term conversational
  memory for {MLLM} agents.
\newblock \emph{arXiv preprint arXiv:2601.03515}, 2026.

\bibitem[Chhikara et~al.(2025)Chhikara, Khant, Aryan, Singh, and
  Yadav]{chhikara2025mem0}
Prateek Chhikara, Dev Khant, Saket Aryan, Taranjeet Singh, and Deshraj Yadav.
\newblock {Mem0}: Building production-ready {AI} agents with scalable long-term
  memory.
\newblock \emph{arXiv preprint arXiv:2504.19413}, 2025.

\bibitem[Cormack et~al.(2009)Cormack, Clarke, and Buettcher]{cormack2009rrf}
Gordon~V. Cormack, Charles L.~A. Clarke, and Stefan Buettcher.
\newblock Reciprocal rank fusion outperforms condorcet and individual rank
  learning methods.
\newblock In \emph{Proceedings of the 32nd International ACM SIGIR Conference
  on Research and Development in Information Retrieval}, pages 758--759, 2009.

\bibitem[Du(2026)]{du2026memorysurvey}
Pengfei Du.
\newblock Memory for autonomous {LLM} agents: Mechanisms, evaluation, and
  emerging frontiers.
\newblock \emph{arXiv preprint arXiv:2603.07670}, 2026.

\bibitem[Ebbinghaus(1913)]{ebbinghaus1913memory}
Hermann Ebbinghaus.
\newblock \emph{Memory: A Contribution to Experimental Psychology}.
\newblock Teachers College, Columbia University, 1913.
\newblock Original work published 1885.

\bibitem[Guti{\'e}rrez et~al.(2024)Guti{\'e}rrez, Shu, Gu, Yasunaga, and
  Su]{gutierrez2024hipporag}
Bernal~Jim{\'e}nez Guti{\'e}rrez, Yiheng Shu, Yu~Gu, Michihiro Yasunaga, and
  Yu~Su.
\newblock {HippoRAG}: Neurobiologically inspired long-term memory for large
  language models.
\newblock In \emph{Advances in Neural Information Processing Systems
  (NeurIPS)}, 2024.

\bibitem[Guti{\'e}rrez et~al.(2025)Guti{\'e}rrez, Shu, Qi, Zhou, and
  Su]{gutierrez2025hipporag2}
Bernal~Jim{\'e}nez Guti{\'e}rrez, Yiheng Shu, Weijian Qi, Sizhe Zhou, and
  Yu~Su.
\newblock From {RAG} to memory: Non-parametric continual learning for large
  language models.
\newblock In \emph{International Conference on Machine Learning (ICML)}, 2025.

\bibitem[Kahneman(2011)]{kahneman2011thinking}
Daniel Kahneman.
\newblock \emph{Thinking, Fast and Slow}.
\newblock Farrar, Straus and Giroux, 2011.

\bibitem[Kang et~al.(2025)Kang, Ji, Zhao, and Bai]{kang2025memoryos}
Jiazheng Kang, Mingming Ji, Zhe Zhao, and Ting Bai.
\newblock Memory {OS} of {AI} agent.
\newblock In \emph{Proceedings of the 2025 Conference on Empirical Methods in
  Natural Language Processing (EMNLP)}, 2025.

\bibitem[Karpukhin et~al.(2020)Karpukhin, O{\u{g}}uz, Min, Lewis, Wu, Edunov,
  Chen, and Yih]{karpukhin2020dpr}
Vladimir Karpukhin, Barlas O{\u{g}}uz, Sewon Min, Patrick Lewis, Ledell Wu,
  Sergey Edunov, Danqi Chen, and Wen-tau Yih.
\newblock Dense passage retrieval for open-domain question answering.
\newblock In \emph{Proceedings of the 2020 Conference on Empirical Methods in
  Natural Language Processing (EMNLP)}, pages 6769--6781, 2020.

\bibitem[Lewis et~al.(2020)Lewis, Perez, Piktus, Petroni, Karpukhin, Goyal,
  K{\"u}ttler, Lewis, Yih, Rockt{\"a}schel, Riedel, and Kiela]{lewis2020rag}
Patrick Lewis, Ethan Perez, Aleksandra Piktus, Fabio Petroni, Vladimir
  Karpukhin, Naman Goyal, Heinrich K{\"u}ttler, Mike Lewis, Wen-tau Yih, Tim
  Rockt{\"a}schel, Sebastian Riedel, and Douwe Kiela.
\newblock Retrieval-augmented generation for knowledge-intensive {NLP} tasks.
\newblock In \emph{Advances in Neural Information Processing Systems
  (NeurIPS)}, 2020.

\bibitem[Liu et~al.(2024)Liu, Lin, Hewitt, Paranjape, Bevilacqua, Petroni, and
  Liang]{liu2024lost}
Nelson~F. Liu, Kevin Lin, John Hewitt, Ashwin Paranjape, Michele Bevilacqua,
  Fabio Petroni, and Percy Liang.
\newblock Lost in the middle: How language models use long contexts.
\newblock \emph{Transactions of the Association for Computational Linguistics
  (TACL)}, 12:\penalty0 157--173, 2024.

\bibitem[Maharana et~al.(2024)Maharana, Lee, Tulyakov, Bansal, Barbieri, and
  Fang]{maharana2024locomo}
Adyasha Maharana, Dong-Ho Lee, Sergey Tulyakov, Mohit Bansal, Francesco
  Barbieri, and Yuwei Fang.
\newblock Evaluating very long-term conversational memory of {LLM} agents.
\newblock In \emph{Proceedings of the 62nd Annual Meeting of the Association
  for Computational Linguistics (ACL)}, 2024.

\bibitem[Packer et~al.(2023)Packer, Wooders, Lin, Fang, Patil, Stoica, and
  Gonzalez]{packer2023memgpt}
Charles Packer, Sarah Wooders, Kevin Lin, Vivian Fang, Shishir~G. Patil, Ion
  Stoica, and Joseph~E. Gonzalez.
\newblock {MemGPT}: Towards {LLMs} as operating systems.
\newblock \emph{arXiv preprint arXiv:2310.08560}, 2023.

\bibitem[Park et~al.(2023)Park, O'Brien, Cai, Morris, Liang, and
  Bernstein]{park2023generative}
Joon~Sung Park, Joseph~C. O'Brien, Carrie~J. Cai, Meredith~Ringel Morris, Percy
  Liang, and Michael~S. Bernstein.
\newblock Generative agents: Interactive simulacra of human behavior.
\newblock In \emph{Proceedings of the 36th Annual ACM Symposium on User
  Interface Software and Technology (UIST)}, 2023.

\bibitem[Rasmussen et~al.(2025)Rasmussen, Paliychuk, Beauvais, Ryan, and
  Chalef]{rasmussen2025zep}
Preston Rasmussen, Pavlo Paliychuk, Travis Beauvais, Jack Ryan, and Daniel
  Chalef.
\newblock {Zep}: A temporal knowledge graph architecture for agent memory.
\newblock \emph{arXiv preprint arXiv:2501.13956}, 2025.

\bibitem[Robertson and Zaragoza(2009)]{robertson2009bm25}
Stephen Robertson and Hugo Zaragoza.
\newblock The probabilistic relevance framework: {BM25} and beyond.
\newblock \emph{Foundations and Trends in Information Retrieval}, 3\penalty0
  (4):\penalty0 333--389, 2009.

\bibitem[Wu et~al.(2025)Wu, Wang, Yu, Zhang, Chang, and Yu]{wu2025longmemeval}
Di~Wu, Hongwei Wang, Wenhao Yu, Yuwei Zhang, Kai-Wei Chang, and Dong Yu.
\newblock {LongMemEval}: Benchmarking chat assistants on long-term interactive
  memory.
\newblock \emph{International Conference on Learning Representations (ICLR)},
  2025.

\bibitem[Wu et~al.(2026)Wu, Zhang, Lin, Xu, Xu, Chen, Zou, Chen, Zhang, Liu,
  Yu, and Wang]{wu2026gam}
Zhaofen Wu, Hanrong Zhang, Fulin Lin, Wujiang Xu, Xinran Xu, Yankai Chen,
  Henry~Peng Zou, Shaowen Chen, Weizhi Zhang, Xue Liu, Philip~S. Yu, and
  Hongwei Wang.
\newblock {GAM}: Hierarchical graph-based agentic memory for {LLM} agents.
\newblock \emph{arXiv preprint arXiv:2604.12285}, 2026.

\bibitem[Xiao et~al.(2024)Xiao, Liu, Zhang, Muennighoff, Lian, and
  Nie]{xiao2024cpack}
Shitao Xiao, Zheng Liu, Peitian Zhang, Niklas Muennighoff, Defu Lian, and
  Jian-Yun Nie.
\newblock {C-Pack}: Packed resources for general chinese embeddings.
\newblock In \emph{Proceedings of the 47th International ACM SIGIR Conference
  on Research and Development in Information Retrieval (SIGIR)}, 2024.
\newblock Introduces the BGE embedding models, incl.\
  \texttt{bge-small-en-v1.5}; arXiv:2309.07597.

\bibitem[Xu et~al.(2025)Xu, Liang, Mei, Gao, Tan, and Zhang]{xu2025amem}
Wujiang Xu, Zujie Liang, Kai Mei, Hang Gao, Juntao Tan, and Yongfeng Zhang.
\newblock {A-MEM}: Agentic memory for {LLM} agents.
\newblock \emph{arXiv preprint arXiv:2502.12110}, 2025.

\end{thebibliography}

\appendix

\section{Prompts}
\label{app:prompts}

For full reproducibility we reproduce the exact prompts the harness uses, verbatim from the repository
(non-ASCII characters are normalised for typesetting). The \emph{answerer} system prompt (used with
\texttt{--reasoning}; it instructs evidence aggregation, most-recent-wins on conflicts, and abstention
only as a last resort):

\begin{lstlisting}[style=prompt]
You answer the user's question using the provided dated context. The context has two parts: a FACTS
index (a digest) AND the full dated CONVERSATIONS below it. The answer is USUALLY present -- search BOTH
parts thoroughly before concluding anything. The dates are real; use them for temporal reasoning.

DIG FOR THE ANSWER FIRST. Most questions ARE answerable from the history -- your job is to find the
evidence, not to give up. If the FACTS digest doesn't contain it, scan the full CONVERSATIONS below.

WHEN THE QUESTION NEEDS MULTIPLE EVIDENCE PIECES (counting, summing, ordering, date arithmetic, duration,
earliest/latest/most-recent, multi-step lookup, knowledge updated over time):
  EVIDENCE: list EVERY relevant dated item from the context, one per line with its date. Be EXHAUSTIVE
  -- scan the whole history; a missed item makes a count or total wrong. Re-read before finalizing.
  REASON: count distinct items / sum / sort by date / compute the date difference / pick the most recent.
  Show the work in 1-2 lines. For durations compute end_date - start_date explicitly.
  ANSWER: <a single concise final answer -- no reasoning on this line>

FOR DATE/TIME/NUMBER ANSWERS: give the MOST SPECIFIC value the context supports (exact date > month+year
> year; exact duration). Read the exact figure from the text -- don't round (e.g. 27 minutes 45 seconds,
not 28 minutes).

WHEN THE QUESTION ASKS FOR A RECOMMENDATION / PREFERENCE: do NOT refuse, do NOT ask a clarifying question,
do NOT give generic advice. Ground the answer in the user's OWN stated history: name the SPECIFIC people,
places, brands, tools, experiences or constraints they mentioned, and tailor the suggestion to those.

FOR SIMPLE SINGLE-FACT QUESTIONS: go straight to 'ANSWER: <fact>'.

CURRENT-STATE questions ('what is my current/latest X', 'where do I work now'): report ONLY the most
recent value by date and explicitly disregard older, superseded ones. The newest dated statement wins.
When facts conflict, ALWAYS trust the most recent one by date.

ABSTENTION -- a careful LAST resort, only after searching the ENTIRE history (facts AND all conversations)
and the SPECIFIC thing asked is genuinely never stated. Do NOT abstain just because it wasn't in the FACTS
digest. If the question presupposes something the user truly never mentioned, reply exactly 'ANSWER: I
don't know' rather than guessing. Never fabricate a value. The 'ANSWER:' line is REQUIRED.
\end{lstlisting}

\noindent The answer template (the context is the only thing that varies across systems):
\begin{lstlisting}[style=prompt]
Today's date: {qdate}

{context}

Question: {question}
Answer:
\end{lstlisting}

\noindent The System-2 fact-extractor system prompt (a non-English worked example is elided for
typesetting; values are kept in the user's original language, only the predicate is English snake\_case):
\begin{lstlisting}[style=prompt]
You are a precise information-extraction engine for a long-term memory system. From a multi-turn
conversation, extract the atomic, durable facts it states about the user and the people/things they
mention (identities, attributes, preferences, relationships, possessions, goals/plans, and events with
their times). Output ONLY a JSON array of objects, each with keys "subject", "predicate", "object",
"text". Use short snake_case predicates (e.g. works_at, lives_in, favorite_color, owns, married_to,
born_in, visited). Capture PREFERENCES explicitly with predicates like likes, dislikes, prefers, avoids,
allergic_to, favorite_<thing>; for each preference or dislike stated, output a SEPARATE fact. Resolve
first-person ('I','my','me') to the user's name when known, otherwise to "user". Capture a stated name as
{"subject":"user","predicate":"name","object":"<Name>"}. LANGUAGE: keep subject and object VALUES (names,
places, brands, free text) in the SAME language the user used -- do NOT translate them; only the predicate
stays English snake_case. "text" is a natural one-sentence statement in the conversation's language. Do
NOT infer or invent facts that are not stated. If there are no durable facts, output [].
\end{lstlisting}

\noindent We grade with the \emph{official} LongMemEval category-specific judge prompts, reproduced
verbatim so scores are leaderboard-comparable:
\begin{lstlisting}[style=prompt]
[single-session-user / single-session-assistant / multi-session]
I will give you a question, a correct answer, and a response from a model. Please answer yes if the
response contains the correct answer. Otherwise, answer no. If the response is equivalent to the correct
answer or contains all the intermediate steps to get the correct answer, you should also answer yes. If
the response only contains a subset of the information required by the answer, answer no.
Question: {q}   Correct Answer: {a}   Model Response: {r}
Is the model response correct? Answer yes or no only.

[temporal-reasoning]  (as above, plus:)
... do not penalize off-by-one errors for the number of days. If the question asks for the number of
days/weeks/months and the model makes off-by-one errors (e.g., predicting 19 days when the answer is 18),
the model's response is still correct.

[knowledge-update]  (as above, plus:)
... If the response contains some previous information along with an updated answer, the response should
be considered correct as long as the updated answer is the required answer.

[single-session-preference]
I will give you a question, a rubric for desired personalized response, and a response. Answer yes if the
response satisfies the desired response. The model need not reflect all points in the rubric; the response
is correct as long as it recalls and utilizes the user's personal information correctly.

[abstention / unanswerable]
I will give you an unanswerable question, an explanation, and a response. Answer yes if the model correctly
identifies the question as unanswerable (it may say the information is incomplete, or give other
information but not the asked information).
\end{lstlisting}

\section{Qualitative Examples}
\label{app:examples}

Table~\ref{tab:qual} shows representative cases drawn from the committed 500-question logs where \lean{}
answers correctly and the full-context baseline does not. Two failure modes recur. \textbf{(i) Lost in the
middle}~\citep{liu2024lost}: the evidence \emph{is} present in full-context's $\sim$79k-token window, yet
it returns ``I don't know,'' while the lean $\sim$9.6k-token slice surfaces it. \textbf{(ii) Stale values
on knowledge-update}: full-context returns an incorrect, non-current value (``30 dozen,'' ``26 minutes and
30 seconds'') while the lean bi-temporal slice---most-recent-wins---returns the current one. These are
illustrative, not cherry-picked headline numbers; all 500 per-question records (prediction, gold,
correctness) are in the repository.

\begin{table}[h]
\centering
\footnotesize
\caption{Representative cases from the \textsc{LongMemEval}\textsubscript{S} logs where \lean{} is correct
and full-context is wrong (ID = the benchmark's question id; predictions verbatim, lightly truncated).}
\label{tab:qual}
\begin{tabular}{@{}l p{2.0cm} p{1.9cm} p{2.7cm} p{2.7cm}@{}}
\toprule
ID & Category & Gold & \engram{} (\lean{}) & full-context \\
\midrule
\texttt{ed4ddc30} & knowledge-update & 20 & 20 dozen & \emph{30 dozen} \\
\texttt{e66b632c} & knowledge-update & 27 minutes and 45 seconds & 27 minutes 45 seconds & \emph{26 minutes and 30 seconds} \\
\texttt{0db4c65d} & temporal-reasoning & 18 days & 18 days & \emph{6 days} \\
\texttt{099778bb} & multi-session & 20\% & 20\% & \emph{I don't know} \\
\texttt{2311e44b} & multi-session & 190 & 190 pages & \emph{I don't know} \\
\texttt{8ebdbe50} & single-session-user & Data Science & Data Science certification & \emph{I don't know} \\
\bottomrule
\end{tabular}
\end{table}

\end{document}